\documentclass{article}

\usepackage{subcaption}

\usepackage{algorithm2e}

\usepackage[english]{babel}
\usepackage{amsfonts}

\usepackage[letterpaper,top=2cm,bottom=2cm,left=3cm,right=3cm,marginparwidth=1.75cm]{geometry}

\usepackage{amsmath}
\usepackage{graphicx}
\usepackage[colorlinks=true, allcolors=blue]{hyperref}

\title{Equalization and Brightness Mapping Modes of Color-to-Gray Projection Operators}
\author{Diego Frias}

\begin{document}
\maketitle

\begin{abstract}
In this article the conversion of color RGB images to grayscale is covered by characterizing the mathematical/algorithmic operators used to project 3 color channels to a single one. Based on the fact that most operators assign each of the $256^3$ colors a single gray level, ranging from 0 to 255, they are clustering algorithms that distribute the color population into 256 clusters of increasing brightness. Thus, it is possible to visualize the way operators work by plotting the sizes of the clusters and the average brightness of the colors within each cluster when converting a synthetic reference image composed of one pixel of each color. The equalization mode (EQ) introduced in this work focuses on cluster sizes, while the brightness mapping (BM) mode describes the CIE L* luminance distribution per cluster. When evaluating a large number of linear operators, only 3 classes of EQ modes and 2 classes of BM modes were found according to the shape of the characteristic curves. In this way, a 6-class taxonomy was defined for the infinite population of linear operators. Another classification of linear operators was introduced, based on parameters, which was used to scan a wide range of operators during the study of EQ mode patterns. In this classification, each operator is assigned a family according to an unbounded positive parameter calculated by combining its three weights, so that all operators with the same parameter belong to the same family. Members within each family are identified according to one of the weights. The theoretical/methodological framework introduced was applied in a case study considering the equal-weights uniform operator, the NTSC standard operator and an operator chosen as ideal to lighten the faces of black people to improve facial recognition in current biased classifiers. It has been found that the uniform and standard operators have the same type of BM mode while the standard and ideal operators have the same type of EQ mode. In addition, it was found that most current metrics used to assess the quality of color-to-gray conversions better value one of the two classes of BM mode: the regular type. This is because error metrics, in general, favor transformations in which the distances between the luminances of all color pairs in the image and the distances between the luminances assigned to the grayscale pairs are more similar \cite{c2g_metric_2008_bras}. Therefore, operators with BM mode of the regular type, in which there is a high linear correlation between the average luminance of the colors in the clusters and the brightness of the corresponding gray channel of the respective clusters, will have a smaller error measure than the operators of irregular type. This contrasts with the fact that the best operator selected for face lightening has an irregular type BM mode. Therefore, this cautions against using these general metrics for specific purpose color-to-gray conversions. It should be noted that eventual applications of this framework to non-linear operators can give rise to new classes of EQ and BM modes. The main contribution of this article is to provide a tool to better understand color to gray converters in general, even those based on machine learning, within the current trend of better explainability of models.
\end{abstract}

\section{Introduction}






Image decolorization, better known as transforming color images to grayscale, is a process in which a multichannel (polychromatic) image is converted to a single-channel (monochrome) version. Decolorization is a computational data reduction process that produces grayscale images containing the most important information, while saving at least 2/3 of the storage space.

In addition to size reduction, a grayscale image can better display object contours and surface textures, which becomes important in object detection-tracking tasks in computer vision systems.
Decolorization can also be widely applied in the field of image segmentation, compression and visual arts, while it is very common when printing color images in grayscale, in order to save costs and shorten printing times \cite{decolorization_2022}.

The great difficulty of image decolorization is the loss of contrast, which has caused the development of a wide variety of methods of different nature and complexity. Assuming that there is no invariant method that performs optimal decolorization of any type of input image, the current trend focuses on adapting the configuration of methods according to the characteristics of the color image being processed. In this direction, researchers have proposed methods based on local and global features. However, there is a tendency towards increasing complexity and computational cost, which limits the application of most methods in real-world applications. For this reason, several authors have pointed to the importance of resuming research on linear operators, simpler and yet very effective and efficient \cite{rgb_out_2013,opt_ga_2016,review_decolor_2021}.

Linear operators transform colors, generally represented by 3 channels: Red, Green and Blue, denoted by RGB, to grayscale by applying different weights to each color channel. As the weights are real numbers, there are infinite combinations of the 3 weights, which opens the possibility that there is always a combination that makes the decolorization in the most convenient way for the type of image being decolorized.

The difficulty lies in the fact that it is necessary to implement a process to search for the optimal combination of parameters of the linear operator, for which it is necessary to define a way of measuring the quality of the transformation \cite{quality_c2g_2015}. For this, it is first necessary to define the quality criteria, which can change from one application to another. Secondly, it is necessary to define whether the choice of the best operator that meets the established quality criteria will be done subjectively, by human evaluators \cite{cadik_2007}, or objectively by the decolorization algorithm itself \cite{c2g_metric_2008_bras}. 

In the first case, it is necessary to establish an evaluation protocol that guarantees the parsimony of the choice, without perceptive bias and without external influences to the process \cite{cadik_2007}. In the second case, it is necessary to define a metric that represents the established quality criteria, which is not an easy task in very specific applications, such as the one was used as a case study in this article: facial whitening of black people's face images to improve the performance of people recognition algorithms. In general, developing metrics for applications where improvement needs to occur in specific regions of the images is a challenge in all areas, fundamentally in the medical field. Among the difficulties of this problem, it can be cited the need to previously apply image segmentation methods.

Still regarding the choice/definition of metrics, it is necessary to choose whether the metric will be calculated evaluating only the converted image (output) or whether it will be based on the comparison between the output and input images. The vast majority of metrics are of the second type, which considerably increases the computational cost of evaluating the quality of the conversion \cite{c2g_metric_2008_bras, Lu2012, quality_c2g_2014,quality_c2g_2015}. Interestingly, quality metrics are defined based on error measures, so minimizing error maximizes conversion quality. The basis of most of this type of metric is the preservation of the final and initial luminosity difference for each pair of pixels in the original image. The initial luminosity difference is obtained by transforming the pixel colors into the CIEL*a*b* color space, or similar, using the L* dimension as a measure of luminosity and the final one by the difference in gray levels assigned to each color. As the computational complexity of these metrics depends polynomially on the number of pixels, it is common to select only a few colors, called quantized colors. Color quantization is in fact a clustering method with the quantized colors being the centroids of clusters or the average of the colors in each cluster. Even though typically the three colors R, G, and B are used, another popular choice is the CIEL*a*b* color space, in which Euclidean distance correlates best with perceptual color difference.

It is worth mentioning that this type of error is minimized by a color-to-gray projection operator that uses the formula 255 L*/100 to generate grayscale images. However, this type of operator, which minimizes the error thus defined, is not recommended in images with a profusion of isoluminating colors.

Continuing with the challenges of the optimal operator search problem, having established the mechanism for evaluating the quality of the converted images, it is necessary to implement the search algorithm. There are two approaches for this: exhaustive finite grid search or gradient-oriented search. Neural models, increasingly popular, belong to the latter approach.

Now, it is intriguing that, despite the fact that this subject has been intensively researched since the 70s of the last century, that is, it has been studied for more than 50 years, it has not been addressed, the most natural way to analyze the decolorization operators , that is, the qualitative and quantitative characterization of the color mapping performed by the operator. In other words, why hasn't a mathematical framework been established that allows answering the following basic questions:

(1) How are the colors distributed among the different gray levels?,
which can be divided into:
 
(1.a) How many colors are mapped to a given gray level?,

(1.b) What kind of color (with what luminance) is mapped to a given gray level,

(2) How many types of operators are there according to the amount of colors that map to the different levels of gray?

(3) How many types of operators are there according to the luminance of the colors that map to the different levels of gray?

In this work, it has been described a mathematical framework and answer all these basic questions in the context of linear operators. However, this does not limit the application of the framework to non-linear operators.

The answer to question 1.a is called Equalization (EQ) mode and to question 1.b is called Brightness Mapping (BM) mode. The answer to question 3 is given by showing 3 classes and two subclasses of linear operators according to the EQ mode pattern. The answer to question 4 is given by showing 2 classes of BM modes patterns for linear operators.

In order to calculate the EQ and BM modes, the framework includes the concept of reference image that constitutes a complete collection of $256^3$ colors in RGB format and that can be viewed in one and two dimensions. The algorithms for generating the 1D and 2D reference images are part of the framework.

In parallel with the presented framework, a parametric classification of linear operators according to the combination of their 3 weights is introduced. More specifically, with the three weights of an operator, a single parameter is calculated that defines which of the infinite existing families the operator belongs to, and with one of the weights it is identified which, of the infinite members of this family, the operator represents.

The article is structured as follows: The section \ref{sec:bck}  contains basic concepts about digital images and a brief review of the description and classification of the color-to-gray operators. The  section  \ref{sec:mat_meth} describes the reference image and EQ/BM modes, both qualitatively and mathematically. The section \ref{sec:stydy_case}  describes the case study in this work where the EQ and BM modes of a linear operator selected for a specific color-to-gray task are compared with those of traditional operators. In the section \ref{sec:res_disc} the results are discussed and in the section \ref{sec:concl} conclusions are drawn summarizing the contributions of this work.

\section{Background}\label{sec:bck}

\subsection{Digital Images}

Consider color images $C$ of resolution $H\times V$, where $H$ and $V$ are the number of pixels (color dots) in the horizontal and vertical directions, respectively. The color image has $D>1$ color channels, that is the color of the pixels is a $D$-dimension vector with integer values ranging from $0$ to $L$, where $L=256$ for 8-byte color coding. Vector components represent the intensity of the base colors (channels). The color of the pixel at position $(i,j), i \in [1,H], \ j\in [1, V ]$  is denoted as 
$$C(i,j )= [c_{i,j,1},c_{i,j,2},\dots,c_{i,j,D}]$$ with channel strengths $c_{i,j,d} \in[ 0,L ]$ for $d=1,2,\dots,D$. In the case $D=1$ it is a monochromatic image, also called a grayscale image, while in color images $D=3$ in most color models (RGB, HSV, YCbCr, CIELab, CIELUV, etc.) or $D=4$ in some models as RGBA and CYMK \cite{color_model_review}.

Here it has been considered the RGB color model whose channels are: RED: $d=1$,  coded by $[255,0,0]$, GREEN: $d=2$, coded by $[0,255,0]$, and BLUE: $d=3$, coded by $[0,0,255]$. However, this approach can be extended to other color models. 

\subsection{Transforming RGB images into grayscale}
The grayscale image pixel value $l$ ranging from $0$ to $255$ is proportional to the brightness of the corresponding printed/displayed dot. For this reason it can be assigned a brightness 
\begin{equation}
b(l)=100 \frac{l}{255}
\label{eqn:brightness}
\end{equation}
ranging from $0$ to $100 \%$. Therefore, transforming RGB images to grayscale implies converting colors to brightness levels by reducing the number of color channels from 3 to 1. In this process, obviously, information is lost, preventing the inverse transformation.
Note that since the $256^3=16.777.216$ colors of an RGB image need to be mapped to only $256$ degrees of brightness, multiple RGB colors will be mapped to the same brightness. This mapping is performed with a projection operator denoted by  $P$, which builds a grayscale image $G \in \mathbb{N}^{H  \times V}$ from a $3$-channel image $C \in \mathbb{N}^{H \times V \times 3}$ in the form
\begin{equation}
G=P(C)
\label{eqn:def_P}
\end{equation}

\begin{figure}[h!]
    \centering
    \includegraphics[scale=0.5]{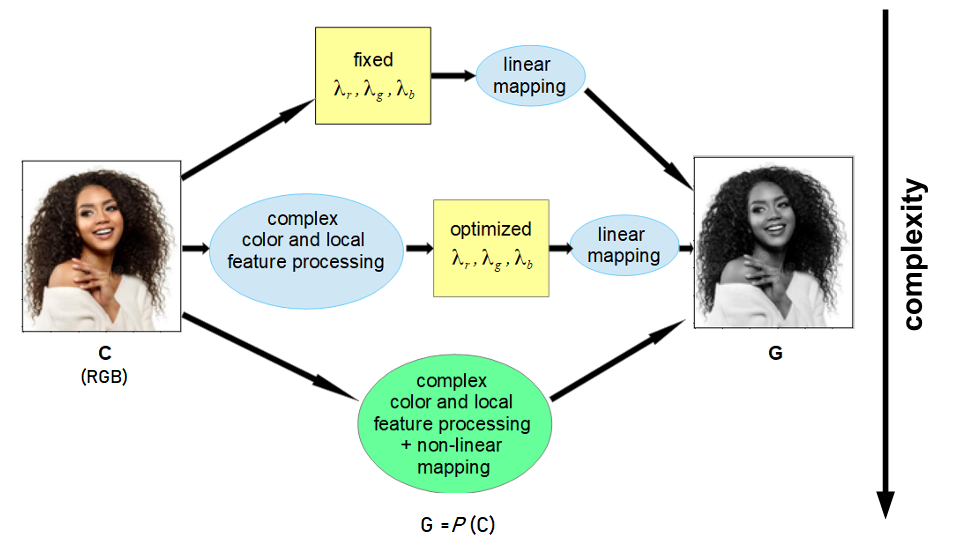}
    \caption{Families of projection operators according to their complexity. }
    \label{fig:pipeline}
\end{figure}

\subsubsection{Classification of Projection Operators}\label{subsubsec:classification}

The projection operators $P$ can be classified within three large families grouped by complexity in figure \ref{fig:pipeline}. Most operators end up with a linear projection using weights for each color channel denoted here as $\lambda_r$ (red ), $\lambda_g$ (green), $\lambda_b$ (blue). In the simplest case the $\lambda$ set is constant \cite{ntsc}, while in the case of moderate complexity the $\lambda$ set is optimized to match the input image properties (global and local color and space features) to the desired output feature (keep or improve contrast, brightness or sharpness) \cite{Lu2012, rgb_out_2013,Liu2015,Liu2017,review_decolor_2021}. The more complex case does not use final linear mapping \cite{bala_2004,gooch_2005,cadik_2007, grunland_2008, non_linear_2009, song_2010,nafchi_2017}. To this family belong machine learning methods in general and neural networks in particular \cite{lin_2008,Hou_2017, learn_2018, zhang_2018,learn_2019}.

Due to the widespread use of linear projection in most operators, there will be described below some of its mathematical and computational properties.

\paragraph{Linear operators:} Let $G={\cal{L}}(C)$ denote a linear transformation that calculates the pixel values of $G$ as 
\begin{equation}
g(i,j)=\text{int} \bigg(\sum_{d=1}^{3} \lambda_d \ c(i,j,d)\bigg) \in \{0,1,\dots,255\}
    \label{eqn:def_linear_op}
\end{equation}
being $\lambda_d \in ]0,1[$\footnote{values $0$ and $1$ are excluded from the valid interval in the model.}\footnote{In \cite{lambda_dif2017} they used $\lambda \in[-1,1]$.}  coefficients that satisfy
\begin{equation}
{\sum_{d=1}^{3} \lambda_d = 1}
    \label{eqn:lambda_constraint}
\end{equation}

In other words, the linear mapping ${\cal{L}}$ is performed by calculating a weighted average of the intensities of the RGB channels using $\lambda$s as weights. It is also common to describe the result of a pixel in grayscale $g(i,j)$ as the dot product of the vector of colors of the pixel in the RGB image by the vector of weights.

The most widely used linear operator is known as the National Television System Committee (NTSC) standard and have weights $\lambda_r=0.299, \ \lambda_g=0.587, \ \lambda_b=0.114$. Another linear operator that can be used as reference is the uniform operator having equal weights $\lambda_r=\lambda_g=\lambda_b=1/3$. 

The computation of the image $G$ with $n=HV$ pixels requires $nD$ products and $n(D-1)$ additions, totaling $n(2D-1)$ operations. Therefore the time complexity of this step is ${\cal{O}}(nD)$.

It is important to note, that despite using three weights, the projection operators are formally functions of two variables, because only two of the three weights can be freely chosen, and the third weight is calculated from  equation \ref{eqn:lambda_constraint}.

\subsection{Current Characterization of Decolorization Algorithms}\label{subsec:curr_characterization}

In a recent work \cite{review_decolor_2021} reviewed 25 state-of-the-art decolorization algorithms. They were classified according to 4 criteria: (1) color space: RGB(15)-Lab(6)-Others(4), (2) Mapping type: global(17)-local(8), (3) method complexity: low(9)-moderate (7)-high(9), and (4) optimization for the input image: yes (13)-no(12).

The majority of algorithms (15/25) operate on RGB images taking into account global properties (17/25), there being a balance between simple (9/25) and very complex (9/25) algorithms and between algorithms that optimize their parameters for each input image (13/25) and those that use fixed parameters (12/25).

Analyzing in more detail the methods that end up with linear mapping, it has been realized that there are two approaches in the way the optimal combination of weights is sought: (1) solving an optimization problem in continuous solution space based on objective function gradient \cite{opt_ga_2016}, and (2) performing an exhaustive search in a discrete solution space. The first work published using this last approach was \cite{Lu2012} who discretized the solution space of $[\lambda_r, \lambda_g, \lambda_b]$ in the interval of $[0, 1]$ with interval $0.1$, so that each weight can take one of the following $11$ values $\{0.0,0.1,0.2,...,1.0\}$. However, due to the constraint \ref{eqn:lambda_constraint}, the search space is reduced to $66$ sets of candidate weights, that is, $66$ linear projection operators to be evaluated according to the chosen performance metric\footnote{In general, the number of candidates for V discrete values is $V(V+1)/2$.}. Thus, the optimization problem is reduced to finding the best solution among a known finite set of candidates, which can be easily done through an exhaustive search. This discrete space search was later used by \cite{rgb_out_2013,Liu2015,Liu2017} and more recently by \cite{review_decolor_2021}.

Unfortunately, the best linear operators found are not reported in any of these publications, being limited only to showing the images produced by them, compared with those obtained with standard mappers, or with other similar methods.

In all cases there is a need to characterize the linear projection operators, in relation to the peculiarities of their mappings. For this reason it has been decided to characterize the operators that are considered candidates in the discrete search space approach described above. This study may provide useful insights into the functional range of tested operators, as well as, probably, associate a certain type of operator with a certain type/nature of images.

\section{Material and Methods}\label{sec:mat_meth}

This section is structured as follows: First a synthetic reference image that has all the colors of the RGB spectrum is presented, followed by the formalization of families of linear projection operators based on a real parameter that is calculated from the weights. Next, the concept of Equalization (EQ) mode is introduced to characterize the output of an operator when the reference image is the input. Finally, the concept of Brightness Mapping (BM) mode is introduced to further characterize the projection operators also using the reference image as input.

\subsection{The Reference Image}

\begin{figure}[h!]
    \centering
    \includegraphics[scale=0.4]{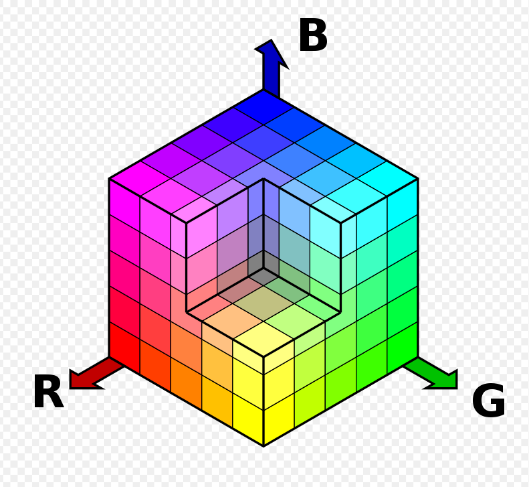}
    \caption{RGB cube \cite{rgb_cube}}
    \label{fig:RGBcube}
\end{figure}

In order to establish a basis for comparing different operators, a synthetic reference image, that is more of a concept than an image in itself, is introduced. What ultimately characterizes this reference image is that it is formed by $L^d$ pixels with the $L^d$ combinations of variables that define the $d$-dimensional color space used. In the case of the most common used RGB color model, $L=256$ and $d=3$, and then the reference image has $256^3$ pixels of different colors, formed by the combinations of $R\in[0,255], G\in[0,255]$ and $B\in[0,255]$. In figure \ref{fig:RGBcube} an illustration of the RGB cube is shown.

\begin{figure}[h!]
    \centering
    \caption{Reference 1D and 2D images composed of $256^3$ pixels of different RGB colors. In the 1D version, the line of pixels was replicated in order to visualize a bar.}
    \includegraphics[scale=0.65]{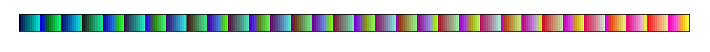}
    \includegraphics[scale=0.12]{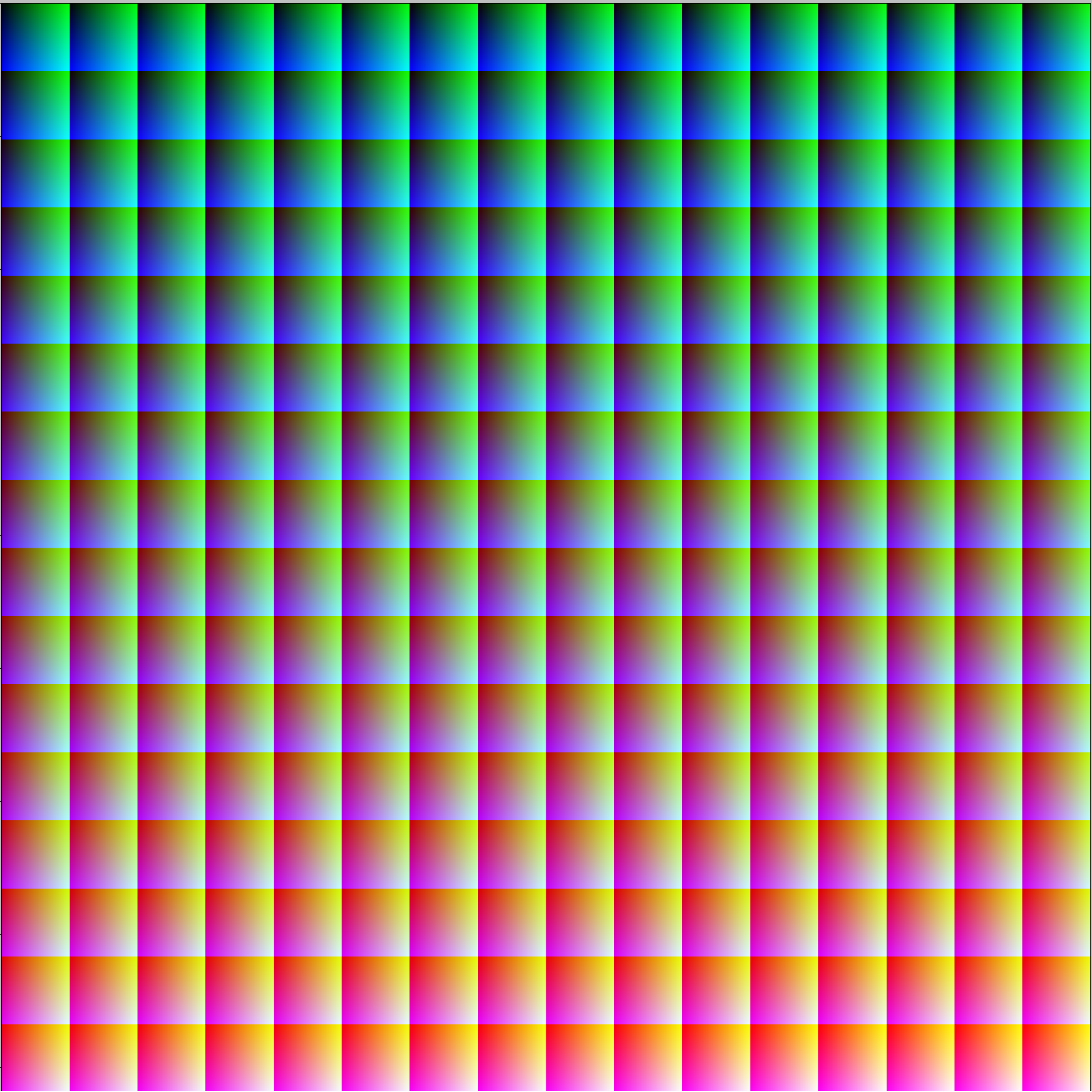}
    \label{fig:ref_1D2Dimage}
\end{figure}

For a practical purpose, the reference image is just a full collection of RGB colors, but it can be visualized and used, for example, to identify the pixels (colors) that were projected to a certain brightness channel. For this reason the reference image is visualized in 1 and 2 dimensions, following a certain ordering of the pixels (colors) as shown in figure \ref{fig:ref_1D2Dimage}. The spatial ordering of colors is determined by algorithms \ref{alg:ref_1D_img} and \ref{alg:ref_2D_img} below (in Python).

\begin{algorithm}[h!]
\caption{Python algorithm for generating the 1D reference image}\label{alg:ref_1D_img}
\KwData{$L=256$}
\KwResult{$C1d$}
{
import $numpy$ as  $np$ \\
$C1d$=[  ] \\
for $r$ in range($L$): \\
$\ \ $ $\ \ $for $g$  in  range($L$): \\
$\ \ $ $\ \ $  $\ \ $      for $b$  in range($L$): \\
$\ \ $$\ \ $$\ \ $ $\ \ $    $\ \ $       $C1d.append([r,g,b])$ \\
$C1d$=np.array($C1d$) \\
$\ $
}
\end{algorithm}

\begin{algorithm}[h!]
\caption{Python algorithm for generating the 2D reference image}
\label{alg:ref_2D_img}
\KwData{$L=256$, $H=16$}
\KwResult{$C2d$}
{
import $numpy$ as  $np$ \\
$V=L // H$ \\ 
$C2d=np.zeros((L*V,L*H,3),dtype=int)$ \\
$for \ r \ in \ range(L): $\\
$\ \ $ $\ \ $ $v=r//H$ \\
$\ \ $ $\ \ $ $ h=r-v*H$ \\
$\ \ $  $\ \ $ $for \ g \ in \ range(L):$ \\
$\ \ $ $\ \ $ $\ \ $ $for \ b \ in \ range(L):$ \\
$\ \ $ $\ \ $ $\ \ $ $\ \ \ $ $C2d[v*L+b,h*L+g] = [r,g,b]$ \\
$\ $
}
\end{algorithm}

While the 1D image is an ordered list of colors, beginning with $[0,0,0]$ (black) and ending with $[255,255,255]$ (white), the 2D reference image is built by dividing the red axis of the RGB color cube into $256$ slices (boxes) of $256 \times 256$ pixels and placed them side by side in an orderly fashion from lowest to highest red intensity in a $16 \times 16$ matrix. Within each box the green intensity increases towards the right and the blue intensity downwards.

\subsection{Defining Families of Linear Operators}

Having that the parameters of the projection operator must satisfy the constraints
\begin{equation}
\sum_{d=1}^{3} \lambda_d=1
    \label{eqn:sum_lambda}
\end{equation}
and 
\begin{equation}
   0<\lambda_d<1, \ \forall d\in[1,D]
    \label{eqn:lim_lambda}
\end{equation}
if an additional constraint is added to the problem, it is possible to calculate one more $\lambda$, remaining just one of them free to define.  
For this purpose two weight ratios can be introduced
$$r_{gr}=\frac{\lambda_g}{\lambda_r}, \ \ r_{bg}=\frac{\lambda_b}{\lambda_g}$$
such that for any linear operator knowing just one of the $\lambda$s and one of such ratios: $r_{gr}$ or $r_{bg}$, the other two $\lambda$s become determined. For example, given $r_{gr}$ and $\lambda_g$ or $\lambda_r$ it can be calculated $\lambda_r$ or $\lambda_g$, respectively, as $\lambda_r=\lambda_g/r_{gr}$ or $\lambda_g=\lambda_r r_{gr}$. Having $\lambda_r$ and $\lambda_g$, it holds that $\lambda_b=1-\lambda_r-\lambda_g$. Similarly can be done knowing $r_{bg}$ and $\lambda_b$ or $\lambda_g$. Therefore, all operators having a similar ratio can be grouped into a family of operators that satisfy a specific constraint. 


A reader may note that since the introduced ratios relate two of the three weights, a third ratio $r_{br}=\lambda_b/\lambda_{r}$ can still be defined, but this is dependent on the other two, as $r_{br} = r_{bg}/r_{gr}$.

Using one of the two reasons introduced to define the projection operator families, there is a problem that an operator can belong to two different families, one according to the ratio $r_{gr}$ and the other to the ratio $r_{ bg} $. That is, the classification of the operator would depend on the ratio used as a reference. For example, operators with weights $[\lambda_r,\lambda_g,\lambda_b]=[0.4, 0.2, 0.4]$ and $[0.2,0.1,0.7]$ have $r_{gr}=0.5$ so they belong to the same family according to $r_{gr}$, but they have $r_{bg}=2$ and $r_{bg}=7$, respectively, which classify them into different families according to $r_{bg}$. To avoid this duality, the ratio $K=r_{gr}/r_{bg}$ is introduced, which is equivalent to

\begin{equation}
    K=\frac{\lambda_g^2}{\lambda_r\lambda_b}
    \label{eqn:def_K}
\end{equation}
where $K\in \mathbb{R}^{+}$ is a parameter that defines the family of the projection operators. This allows us to describe any linear operator by setting just one of $\lambda$s and the family value $K$. Note that each combination of $\lambda$s will produce a unique value of $K$ and also that different combinations of $\lambda$s can produce the same $K$. 

To derive the equations to find the other two $\lambda$s knowing only one $\lambda$ and $K$, let chose $\lambda_b$ as a free parameter, therefore using eq. \ref{eqn:sum_lambda} 
\begin{equation}
\lambda_r= 1-\lambda_g-\lambda_b
    \label{eqn:lr}
\end{equation} 
can be replaced into eq. \ref{eqn:def_K}, that is
\begin{equation}
    \frac{\lambda_g^2}{(1-\lambda_g-\lambda_b)\lambda_b}=K
    \label{eqn:def_K1}
\end{equation}
from where a quadratic equation in $\lambda_g$ is obtained in the form
\begin{equation}
\lambda_g^2+K\lambda_b\lambda_g+K\lambda_b(\lambda_b-1)=0
    \label{eqn:quadratic}
\end{equation}
The two formal solutions are:
\begin{equation}
\lambda_g=-\frac{K\lambda_b}{2}\pm \sqrt{\bigg(\frac{K\lambda_b}{2}\bigg)^2+K\lambda_b\big(1-\lambda_b\big)}
    \label{eqn:2sol_quadratic}
\end{equation}
However, as $\lambda$s are all positive, only the positive radical has sense, and then the unique feasible solution is
\begin{equation}
\lambda_g= \sqrt{\bigg(\frac{K\lambda_b}{2}\bigg)^2+K\lambda_b\big(1-\lambda_b\big)}-\frac{K\lambda_b}{2}
    \label{eqn:1sol_quadratic}
\end{equation}
The equation \ref{eqn:1sol_quadratic} for $\lambda_g$ depends on two real parameters, $0<\lambda_b<1$ and $K>0$. The $K$ parameter naturally defines an infinite family of operators whose $\lambda$ set satisfies eq. \ref{eqn:def_K}, while $\lambda_b$ defines a member of this family.

\subsubsection{Uniform and Standard Operators as Particular Cases}

In section \ref{subsubsec:classification} the uniform and the NTSC standard operators were introduced. Here, their families are identified by calculating the corresponding parameter $K$.

\paragraph{Uniform operator:} Substituting $\lambda_r=\lambda_g=\lambda_b=1/3$ into eq. \ref{eqn:def_K} yields 
$$K=\frac{(1/3)^2}{(1/3) (1/3)}=1$$
Therefore, it is said that the uniform operator is the member for $\lambda_b=1/3$ of the family for $K=1$. Other members of the same family can be defined for different $0<\lambda_b<1$, using eq. \ref{eqn:1sol_quadratic} and eq. \ref{eqn:lr}.

\paragraph{Standard operator:} Substituting $\lambda_r=0.299, \ \lambda_g=0.587, \ \lambda_b=0.114$ into eq. \ref{eqn:def_K} results 
$$K=\frac{(0.587)^2}{(0.299)  (0.114)}=10.109$$

Therefore, it is said that the standard operator belongs to the family for $K= 10.109$ and within this family it is the member for $\lambda_b=0.114$. Other members of the same family can be defined for different $0<\lambda_b<1$, using eq. \ref{eqn:1sol_quadratic} and eq. \ref{eqn:lr}.

In this way, the projection operators were grouped into families according to parameter $K$ and their identifications with $\lambda_b$ as a specific member of their family was provided. 

\subsection{The Equalization Mode of the Projection Operators}

Operators that transform color images into grayscale can be characterized, regardless of their nature, using 2 features: (1) to measure how many colors (not pixels) are mapped to each gray channel, and (2) to describe which colors (not pixels) are mapped to each gray channel. The Equalization mode, denoted by EQ mode, is the first feature. The second feature introduced in this work will be described in the next section.

The EQ mode was defined as the number of colors that are mapped to each gray channel. As in the reference image described above, each pixel corresponds to a different color, the characterization of the operators can always be done based on the reference image. For this reason, to measure the EQ mode, the projection operator $P$ is applied for transforming the 1D reference image $$C1d=\{[0,0,0], [0,0,1],\dots,[255,255,255]\}$$
obtaining a list of gray levels (brightness), that is
$$G=P(C1d)=\{g_0, g_1,\dots,g_{256^3}\}$$ 
where the brightness $g_i\in[0,255], \ i=0,1,\dots,256^3$. The EQ mode of $P$ is given by the $256$-bins histogram of the output list $G$, that is, by the number of occurrences of gray level $j\in[0,255]$, expressed as\footnote{Here the standard notation was used: $\{.\}$ denotes a set, $\big|\{,\}\big|$ denotes the cardinality or number of elements in the set, and $\{a|b\}$ means "a such that b".}
\begin{equation}
EQ_j=\Big|\{ \ g_i, \ i\in[0,256^3] \ | \ g_i==j\}\Big|
    \label{eqn:EQj}
\end{equation}

\begin{figure}[h!]
    \centering
    \includegraphics[scale=0.55]{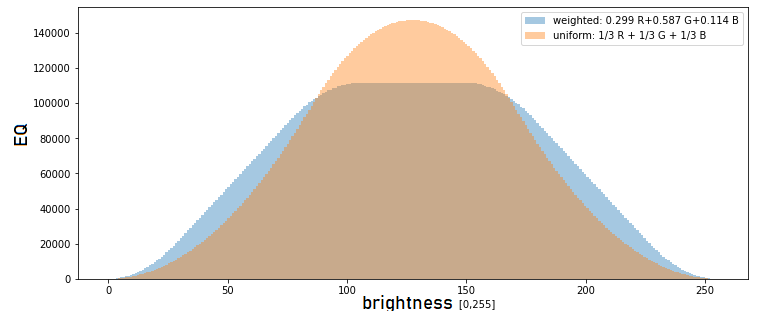}
    \caption{Distribution of RGB colors along the brightness scale for the uniform and standard (weighted) projection operators.}
    \label{fig:grayscales}
\end{figure}

In figure \ref{fig:grayscales} the EQ modes of the uniform and standard operators described above were plotted.

It can be seen that, in both cases, the intermediate brightness receive more pixels than the extreme brightness, and that the standard projector truncates the intermediate brightness (from approximately 85 to 170) by symmetrically distributing the cropped colors to both extremes. The EQ mode reflects the fact that each operator configures a specific equalization, prioritizing some grayscale channels to the detriment of others. 

Even that the EQ mode is a histogram itself, it is not associated with histogram equalization, which is a technique for increasing the contrast of images that have already been transformed to grayscale \cite{histo_eq_2014}.

\subsubsection{Properties of the EQ mode}

The EQ mode is a continuous function $EQ(j)$ of a discrete variable $j\in[0,L-1]$ satisfying 
\begin{equation}
\sum_{j=0}^{L-1} EQ(j) = L^3
    \label{eqn:EQmode_constraint}
\end{equation}
where $L$ is the number of gray levels. Therefore, the priority the equalizer gives to level $j$ is given by 
$$p(j)=EQ(j)/L^3 \in[0,1]$$

\subsubsection{Parametric dependence of the EQ mode}
\begin{figure}[h!]
        \centering
        \begin{subfigure}{.49\textwidth}
        \centering
         \includegraphics[width=0.95\textwidth]{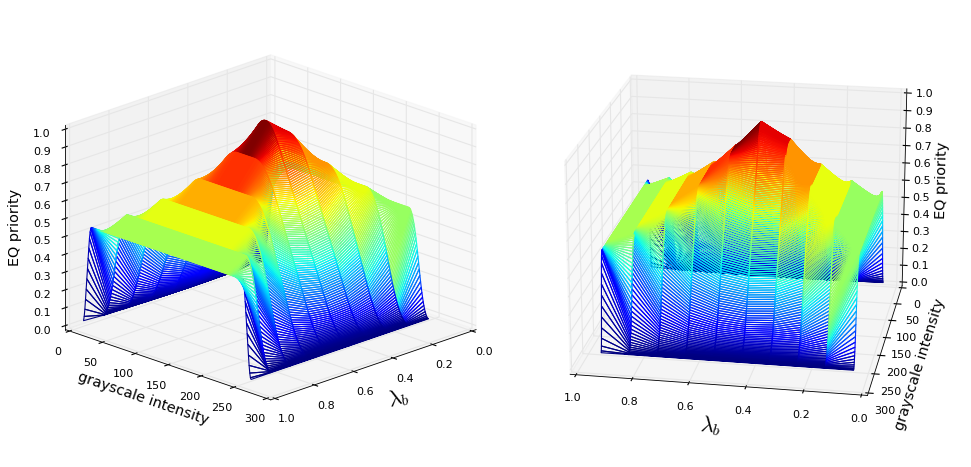}
         \subcaption{$K=0.1$}
        \end{subfigure}
        \begin{subfigure}{.49\textwidth}
        \centering
        \includegraphics[width=0.95\textwidth]{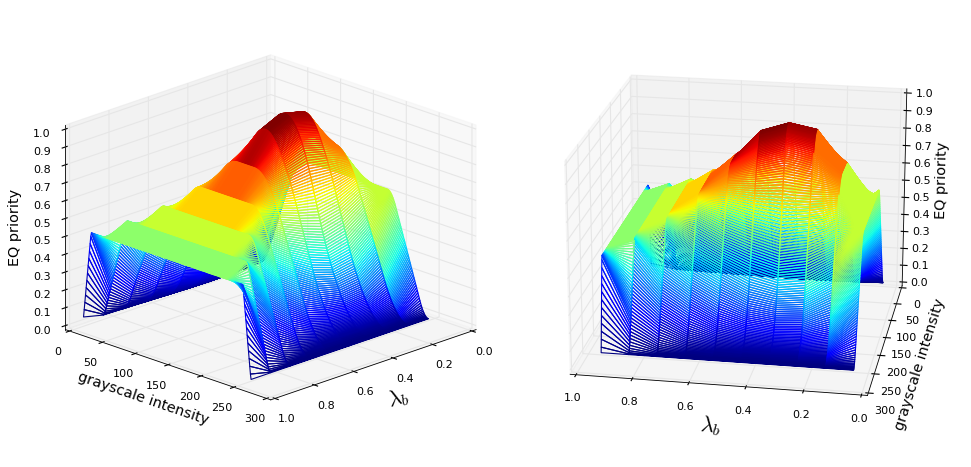}
        \subcaption{$K=1.0$}
        \end{subfigure}
        \begin{subfigure}{.49\textwidth}
        \centering
        \includegraphics[width=0.95\textwidth]{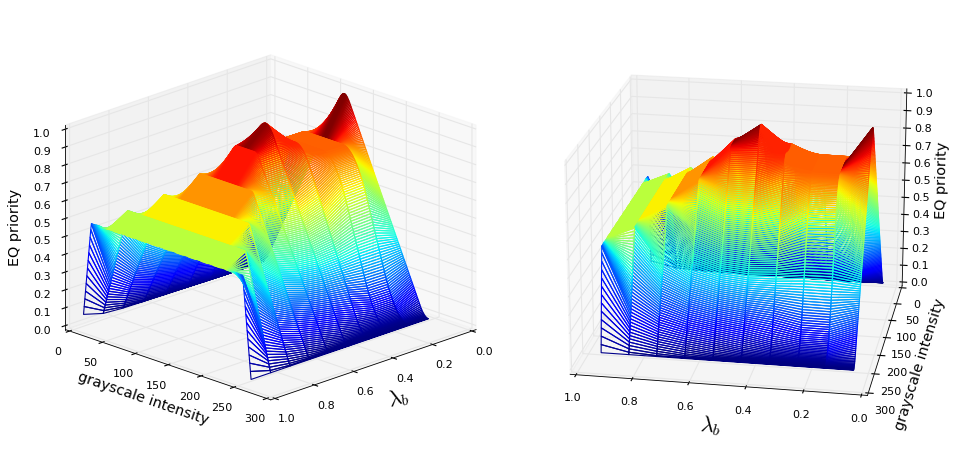}
        \subcaption{$K=10.109$}
        \end{subfigure}
        \begin{subfigure}{.49\textwidth}
        \centering
        \includegraphics[width=0.95\textwidth]{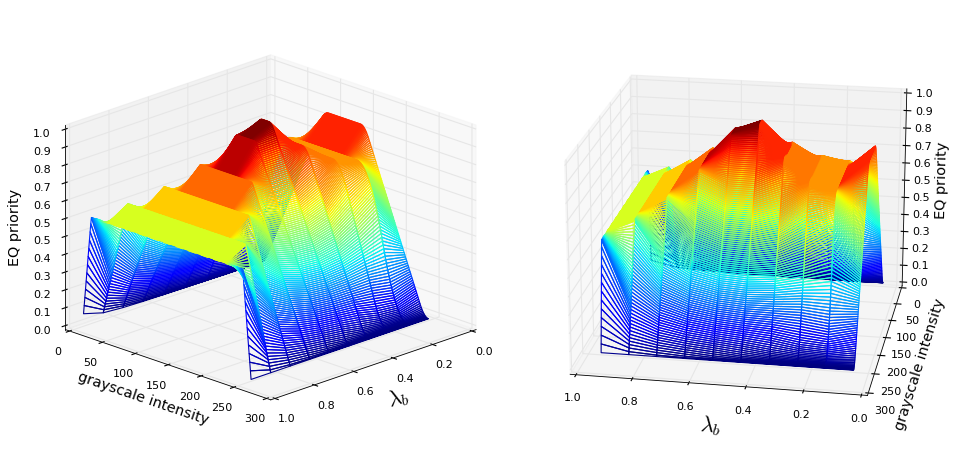}
        \subcaption{$K=20$}
        \end{subfigure}
         \begin{subfigure}{.49\textwidth}
        \centering
        \includegraphics[width=0.95\textwidth]{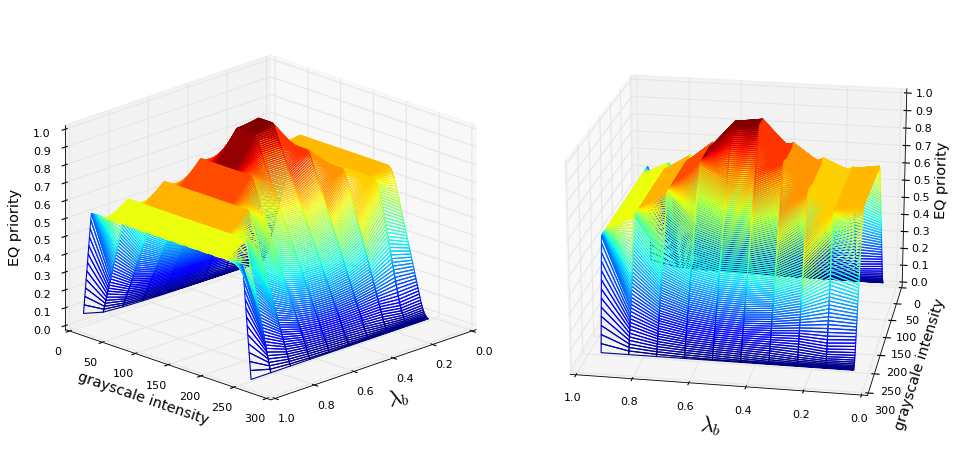}
        \subcaption{$K=60$}
        \end{subfigure}
         \begin{subfigure}{.49\textwidth}
        \centering
        \includegraphics[width=0.95\textwidth]{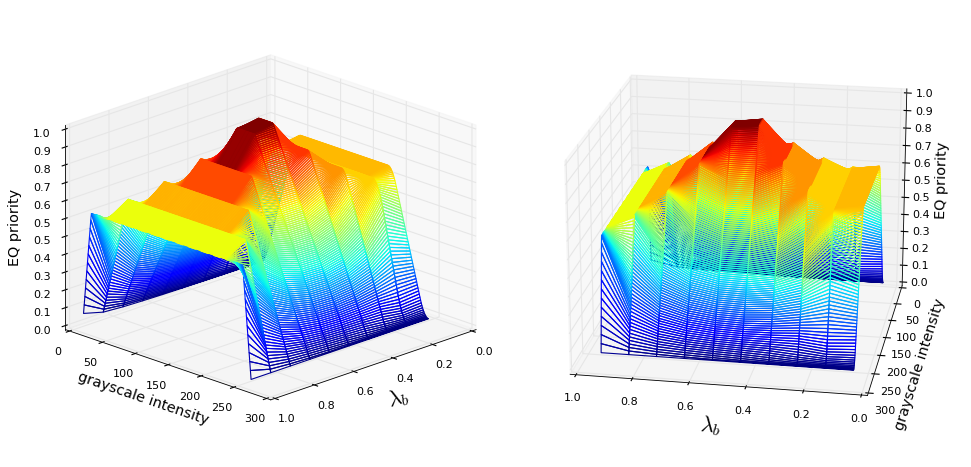}
        \subcaption{$K=100$}
        \end{subfigure}
        \caption{Variation of the EQ mode with $\lambda_b$ for several $K$ values.}
        \label{fig:eq_vslambda_difK}
    \end{figure}

To map the EQ modes in the different families of linear operators, $\lambda_b$ was varied from $0.05$ to $0.95$ with step $0.1$ for some selected families ($K$ values). Figure \ref{fig:eq_vslambda_difK} shows the EQ modes of six operator families,  from top to bottom for $K = 0.1$, $K=1.0$ (uniform operator's family), $K=10.109$ (standard operator's family), and $K=20, \ 60, \ 100$. Two views from different angles were plotted to see the shapes of the EQ modes for low and high $\lambda_b$ values.

Is was noted that:
\begin{enumerate}
    \item There is a global maximum at $\lambda_b=0.5$ for all values of $K$. \item There is a secondary local maximum for $\lambda_b \to 0$ when $K\ge 1$.
    \item There are two minima, one for $\lambda_b \to 0$ and the other for $\lambda_b \to 1$ when $K<1$.
    \item There is a local minimum for $0<\lambda_b<0.5$ when $K\ge 1$, due to the existence of a local maximum at $\lambda_b \to 0$. In these cases the global minimum remains for $\lambda_b \to 1$. 
    \item The $\lambda_b$ of the standard operator ($0.114$) is very close to the local minimum at $\lambda_b=0.2$ of its family for $K=10.109$. 
    \item It appears that the EQ mode converges to an asymptotic form as $K$ increases. Note that the surfaces for $K=60$ and $K=100$ are quite similar.  
    \item It seems that the shape of the EQ mode varies strongly with $\lambda_b$, but not so much with $K$. In order to confirm this assumption in figure \ref{fig:eq_vsK_diflambda} the EQ modes as a function of $K$ were plotted for some values of $\lambda_b$ ranging between $0$ and $1$, more specifically, for $\lambda_b=0.025,\ 0.10,\ 0.25,\ 0.50,\ 0.75$ and $0.95$.  It is important to remember that $K$ identifies the family while $\lambda_b$ identifies the members within the families. 

     \begin{figure}[h!]
        \centering
        \begin{subfigure}{.49\textwidth}
        \centering
         \includegraphics[width=0.95\textwidth]{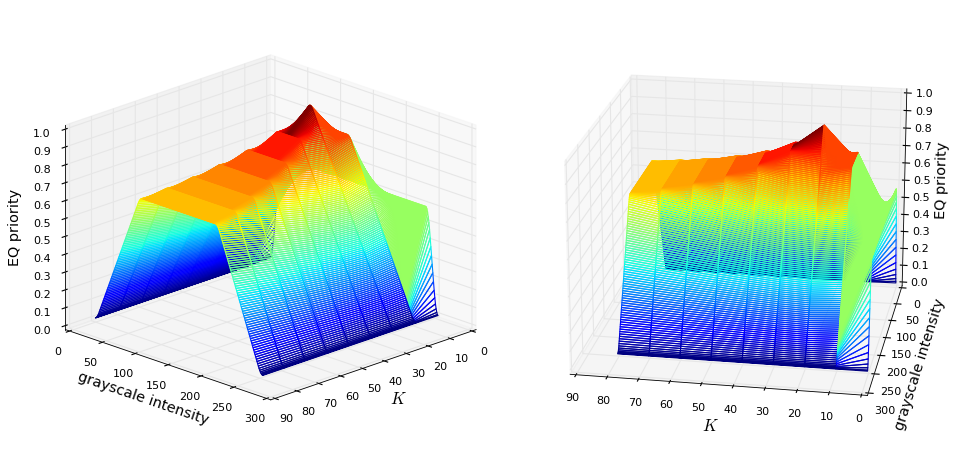}
         \subcaption{$\lambda_b=0.025$}
        \end{subfigure}
        \begin{subfigure}{.49\textwidth}
        \centering
         \includegraphics[width=0.95\textwidth]{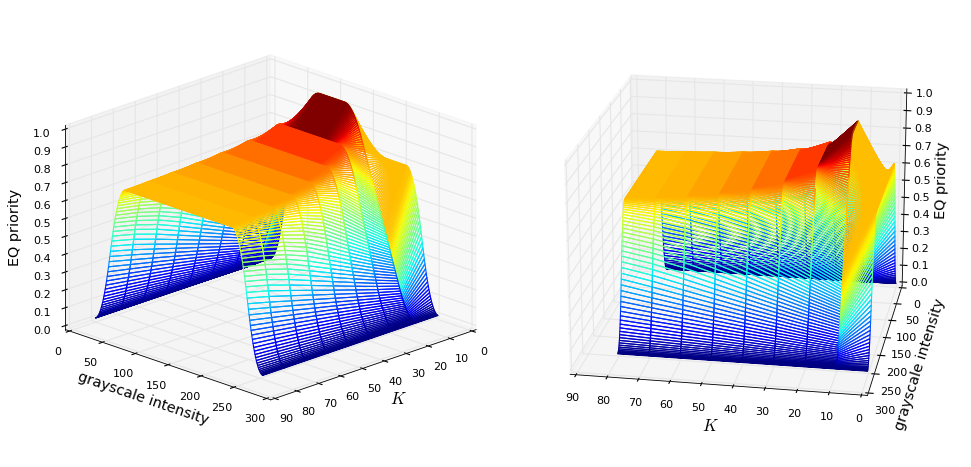}
         \subcaption{$\lambda_b=0.10$}
        \end{subfigure}
        \begin{subfigure}{.49\textwidth}
        \centering
        \includegraphics[width=0.95\textwidth]{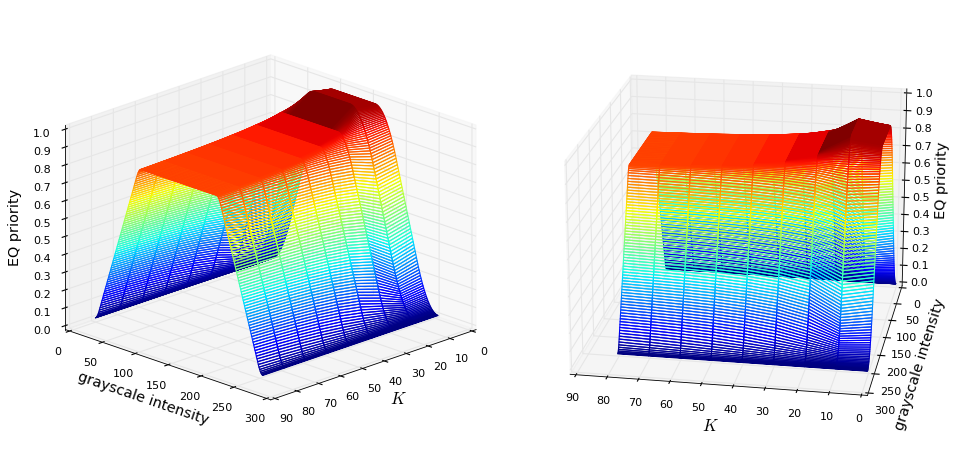}
        \subcaption{$\lambda_b=0.25$}
        \end{subfigure}
        \begin{subfigure}{.49\textwidth}
        \centering
        \includegraphics[width=0.95\textwidth]{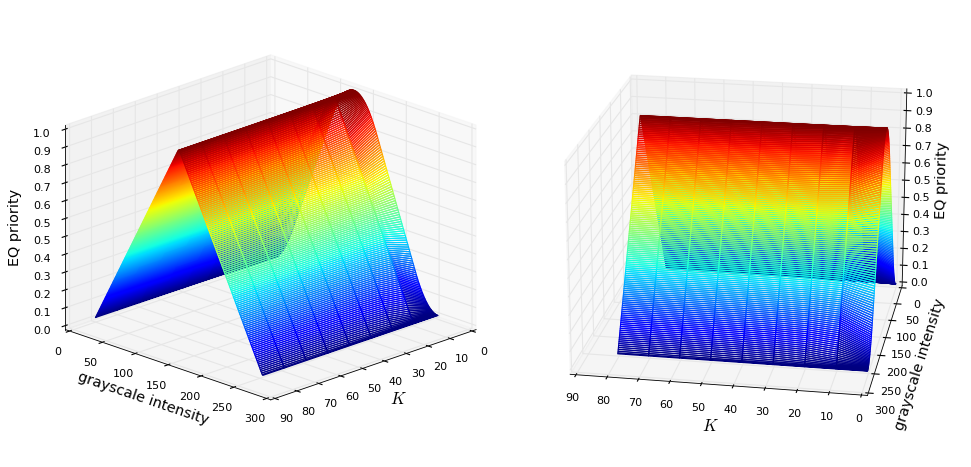}
        \subcaption{$\lambda_b=0.50$}
        \end{subfigure}
        \begin{subfigure}{.49\textwidth}
        \centering
        \includegraphics[width=0.95\textwidth]{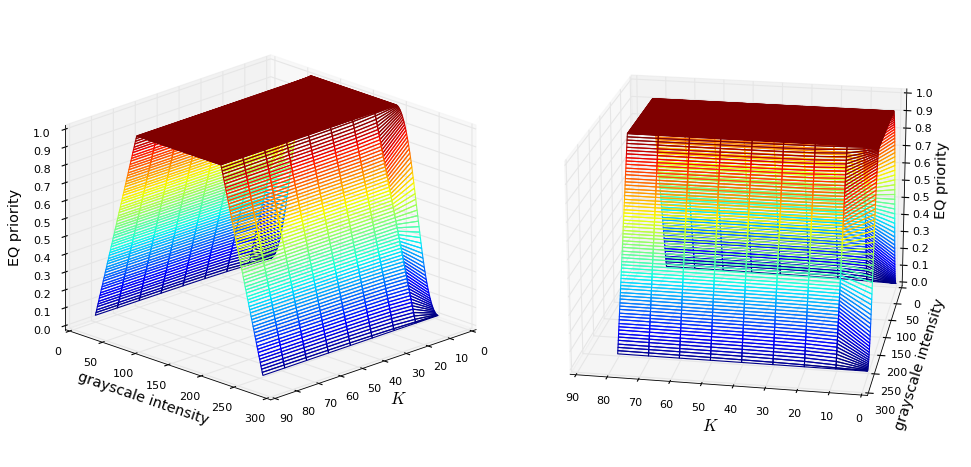}
        \subcaption{$\lambda_b=0.75$}
        \end{subfigure}
        \begin{subfigure}{.49\textwidth}
        \centering
        \includegraphics[width=0.95\textwidth]{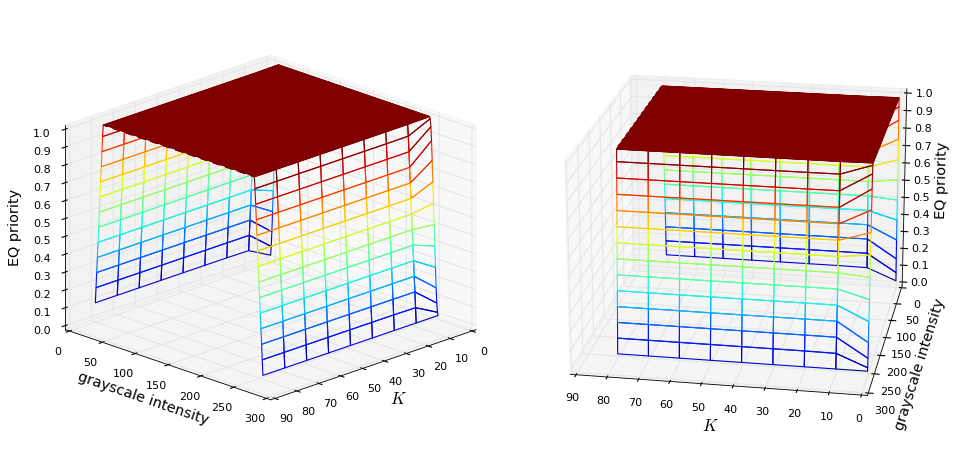}
        \subcaption{$\lambda_b=0.95$}
        \end{subfigure}
        \caption{Variation of the EQ mode with $K$ values (families) for several $\lambda_b$ values.}
        \label{fig:eq_vsK_diflambda}
    \end{figure}

    The results show that, in fact, except for $\lambda_b<0.5$, the EQ mode varies very little from one family to another, that is, varying $K$. For $\lambda_b<0.5$ there is a group of families with a small value of $K\le 20$, in which the EQ mode differs significantly between them. The EQ mode in this range of $\lambda_b$ and $K$ values varies non-linearly between families. It is important to note that the standard operator with $K=10.109$ and $\lambda_b=0.114$ falls in this range.
    
    In this figure it is also easier to identify some specific forms of EQ modes that are described in the next section.
    
\end{enumerate}

\subsubsection{Types of EQ modes}

Regarding the shape of the EQ modes  three classes were identified:
\begin{enumerate}
    \item Bell-shaped, observed for $K\le1$ in the vicinity of $\lambda_b=0.5$. The uniform operator equalization mode belongs to this class.
    \item Trapezoidal, observed, for example, for all $K$ as $\lambda$ increases to $1$. The default operator equalization mode belongs to this class, and 
    \item Triangular, observed, for example, in the family for $K=10.11$ when $\lambda_b \to 0$.
\end{enumerate}

\begin{figure}[h!]
    \centering
    \begin{subfigure}{.99\textwidth}
        \centering
        \includegraphics[width=0.28\textwidth]{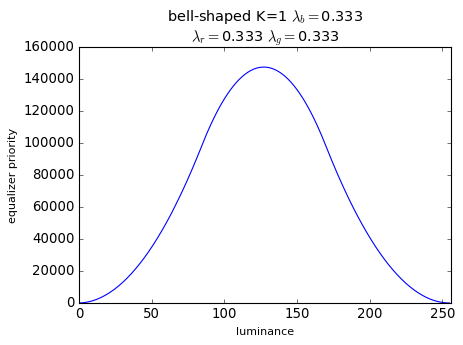}
        \subcaption{}
    \end{subfigure}
     \centering
    \begin{subfigure}{.4\textwidth}
        \centering
        \includegraphics[width=0.7\textwidth]{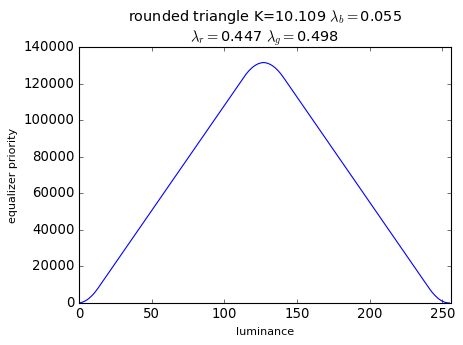}
        \subcaption{}
    \end{subfigure}
     \centering
    \begin{subfigure}{.4\textwidth}
        \centering
        \includegraphics[width=0.7\textwidth]{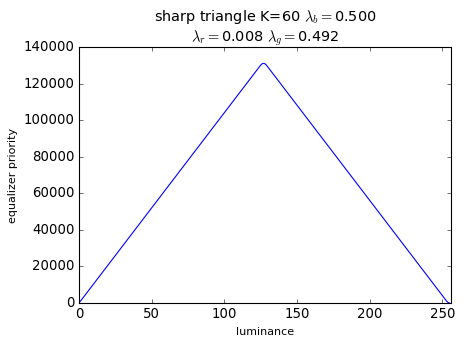}
        \subcaption{}
    \end{subfigure}
     \centering
    \begin{subfigure}{.4\textwidth}
        \centering
        \includegraphics[width=0.7\textwidth]{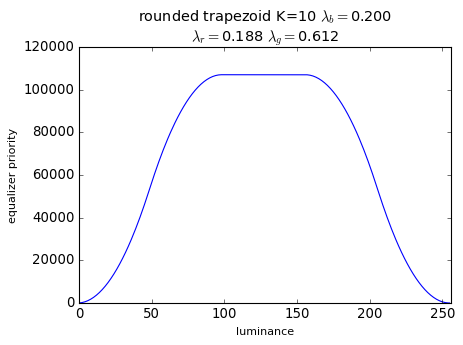}
        \subcaption{}
    \end{subfigure}
        \begin{subfigure}{.4\textwidth}
        \centering
        \includegraphics[width=0.7\textwidth]{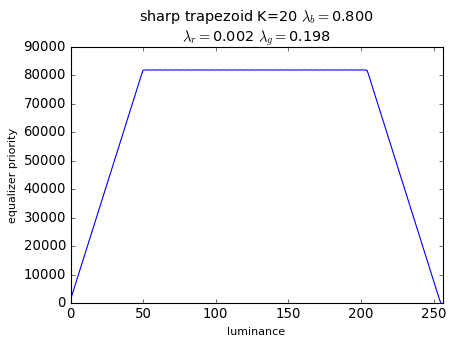}
        \subcaption{}
    \end{subfigure}

    \caption{Types and variants of EQ modes for linear projection operators.}
    \label{fig:equalizer_classes}
\end{figure}

In addition, trapezoidal and triangular shapes have two variants: (1) rounded corners, (2) sharp corners. To illustrate the EQ mode types,  figure \ref{fig:equalizer_classes} shows one example of each class and their variants, indicating the corresponding parameters $K$ and $\lambda_b$, as well as the complementary parameters $\lambda_r$ and $\lambda_g$, for conference.

The sharp EQ modes and especially the triangular EQ mode were somewhat unexpected. In principle, the triangular EQ mode is far from being an attenuator of intermediate gray intensities, it is a symmetrical linear intensifier of that band. What effect this might have on transformed images began to be investigated within a case study described below. 

\newpage
\subsubsection{Remarks on EQ modes}\label{sec:remarks_EQ}

The trapezoidal equalization mode, to which the current standard projection operator belongs, describes a uniform central band attenuator (or plateau type equalizer), which distributes the attenuated colors, symmetrically, to the low and high brightness bands. This redistribution can produce interesting and useful effects, such as the enhancement of the global contrast in the image.

The EQ mode was conceived to characterize operators independently of the input image, so formally it only makes sense in operators that do not adapt to the image being processed. Therefore, adaptive operators, whether to the global characteristics of the image or to the local ones, should not be characterized with EQ mode. Why? In the case of operators based on global characteristics, there will be different EQ modes for each image, which does not prevent the EQ mode from being calculated as described. However, in the case of operators based on local features, the same color can be mapped to different gray levels depending on the local context where the color is found within the same image. This causes the number of projected colors to be greater than $L^3$ because some colors are counted multiple times. However, a dynamically normalized version of EQ mode could still be used in these cases.

Based on what has been discussed here, the EQ mode could be used in the analysis of adaptive operators, allowing to see operator changes for different types of images, contributing to the method's explainability.

\subsection{The Brightness Mapping Mode of Projection Operators}

As the EQ mode only considers the quantitative side of the operator's color distribution, it is natural that operators with similar EQ modes can map very different colors to their channels, causing a different effect on the transformed image. For this reason, it is important to introduce a qualitative operator descriptor, which considers the colors chosen for each channel. The qualitative descriptor must allow, more than characterizing an individual operator, to allow the comparison between operators. In this section a type of qualitative descriptor called Brightness Mapping (BM) mode is described.

\begin{figure}[h!]
    \centering
    \includegraphics[scale=0.5]{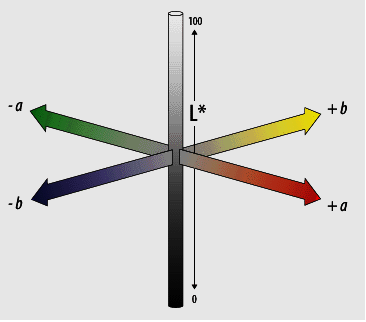}
    \caption{CIEL*a*b* color space \cite{color_model_review}}
    \label{fig:cielab}
\end{figure}

Qualitative descriptors are conceptually designed to extract useful information from color sets that are mapped to different gray channels. Therefore, the main  difficulty in designing this type of descriptor lies in the need to compare color sets of different sizes. One way to approach this problem is to define a metric per channel that does not depend on the number of colors projected onto it, but varies with the population of projected colors. Since each gray channel $j \in[0.255]$ can be associated with a brightness level 
\begin{equation}
    b(j)=100\frac{j}{255} \  \in \ [0\%. \ 100\%], 
    \label{eqn:bj}
\end{equation}
 a metric based on the distribution of the brightness of the colors projected in each channel can be used. The most straightforward way is to use the luminance of the colors based on some reference color space. Here the luminance $L^*$ from the CIEL*a*b color space \cite{color_model_review}, represented in figure \ref{fig:cielab}, was used.  

Consider an operator ${\cal{L}}$ that given a color $c=[R,G,B]$ computes the $L^*={\cal{L}}(c)$, in the following way:

if  $R/255 > 0.04045$  then 
$r = ( \frac{R/255 + 0.055}{1.055} )^{2.4}$
else                   
$r = \frac{R/255}{ 12.92}$

if $G/255 > 0.04045$ then  
$g = ( \frac{G/255 + 0.055}{1.055} )^{2.4}$
else                   
$g = \frac{G/255}{ 12.92}$

if $B/255 > 0.04045$  then
$b = ( \frac{ B/255 + 0.055 }{1.055} )^{2.4}$
else                   
$b = \frac{B/255} { 12.92} \ $. 

With $r,g,b$ calculate 
$Y = 0.2126 \ r +   0.7152 \ g +  0.0722 \ b\ $ and compute the luminance in the form $L^*(c) = 116 \ Y^{1/3}-16$ if $Y > 0.008856$, otherwise as  $L^*(c) = 903.292 \ Y$.

Let $C(j)=\{c_1,c_2,\dots,c_{EQ(j)}\}$ be the set of $EQ(j)$ colors mapped to the gray channel $j$, where $EQ(j)$ is the number of colors mapped to this channel, defined in Eq. \ref{eqn:EQj}. Therefore, ${\cal{L}}(C(j))$ is the set (distribution) of luminances of the colors mapped to the gray channel $j$. The Brightness Mapping (BM) mode introduced in this work is given by the mean of the luminances in all gray channels, denoted as $B(j)=\langle{\cal{L}}(C(j))\rangle, \ j=0,1,\dots,255$, which can be explicitly expressed as
\begin{equation}
    B(j)=\frac{1}{EQ(j)}\sum_{i=1}^{EQ(j)} L^*(c_i) \  \in \ [0\%. \ 100\%], 
\label{eqn:Bj}
\end{equation}

The BM mode is visualized by plotting the calculated brightness $B(j)$ (eq. \ref{eqn:Bj}) as a function of the gray brightness $b(j)$ (eq. \ref{eqn:bj}). That is, it describes the correlation between the actual luminance of the color set $B(j)$ and the brightness assigned by the projection operator to each color set $b(j)$. 

\begin{figure}[h!]
    \centering
    \includegraphics[scale=0.4]{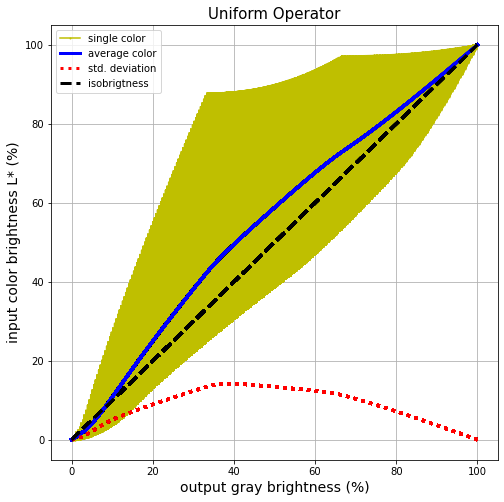}
    \includegraphics[scale=0.4]{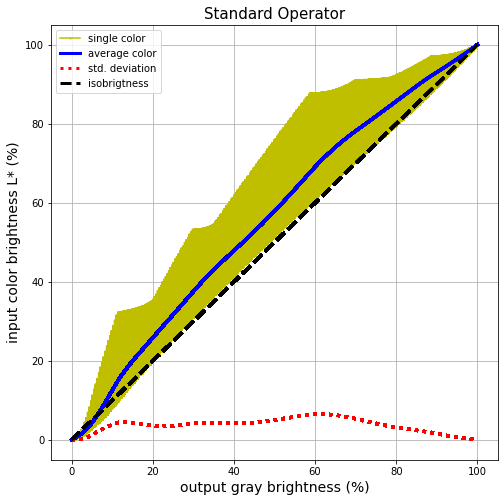}
    \caption{Plot of $B$ vs $b$ for the uniform and standard operators. \underline{Yellow area}: single color brightness cloud, \underline{Blue solid line}: the {\bf{BM mode}} curve = mean brightness $B$ at each gray level. \underline{Black dashed line}: isobrightness function $B=b$, and \underline{Red dotted line}: standard deviation of $B$ at each gray level.}
    \label{fig:BMM_2operators}
\end{figure}

Figure \ref{fig:BMM_2operators} shows the BM modes of the uniform and standard operators analyzed in this work.
To correctly interpret the BM mode, it is needed to highlight that: (1) the two axes vary on the same scale between $0$ and $100\%$, (2) the vertical axis is the average real brightness of the set of colors mapped to each channel, and (3) the horizontal axis is the brightness assigned by the projection operator to the set of colors mapped to each channel. That is, points above the isobrightness $y=x$ line are channels where the operator assigned a luminance lower than the real one, and points below the channels where it assigned a luminance greater than the real one.

Regarding the uniform and standard operators they assign, mostly over the entire range, a brightness moderately lower than the average brightness of the colors projected on each channel, as can be seen from the fact that the BM mode curve is always above the isobrightness line (dashed black line). In other words, these operators produce grayscale images with a moderate loss of luminance relative to the luminance of the reference image.

\subsubsection{Types of BM modes}

In order to analyze the BM modes of a representative sample of linear operators,  $36$ among the $66$ candidate operators that are evaluated in the discrete space search methods \cite{Lu2012}, were selected. The number of operators was reduced, in addition to $66$ being a quite large number to visualize the results, because in  \cite{Lu2012} $\lambda$ values vary in the range $0 - 1$ with step $0.1$, which makes operators map $1$, $2$ or $3$ color channels to grayscale. In other words, when some $\lambda$ is $1$ then only one color is mapped, when some $\lambda$ is $0$ then only $2$ colors are mapped, so $3$ channels are only mapped when none of the $\lambda$s is $0$ or $1$. As in this work only operators that map the $3$ channels are considered, the range was limited to $0.1-0.8$ keeping the step of $0.1$, which results in $8$ values per channel and then $8(8+1)/2=36$ combinations of $\lambda$s.

\begin{figure}[h!]
    \centering
    \includegraphics[scale=1.2]{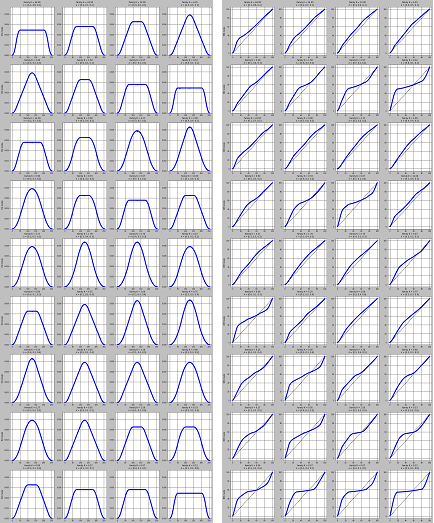}
    \caption{EQ modes on the left and BM modes on the right of 36 linear operators.}
    \label{fig:EQ_BM_modes_36}
\end{figure}

Figure \ref{fig:EQ_BM_modes_36} shows on the right the BM modes of the $36$ considered linear operators. On the left, its EQ modes were added for a better association of the two features.

It can be seen that the BM modes can be classified into only 2 classes called as regular and irregular. The regular pattern (e.g. the 4 operators in the first row) produces systematic smooth subluminance in the sense of assigning gray luminances slightly lower than the average luminance of the colors belonging to almost all gray channels, with the exception of the initial channels ($<\approx. 10\%$ of the grayscale range) where very light overluminance occurs, that is, where the assigned luminance is slightly higher than the average luminance of the colors projected on these gray channels. The irregular pattern (e.g. the 4 operators in the last row) produces variable more intense subluminance in approximately $60\%$ of the grayscale range, followed by a variable intensity overluminance effect in the final $40\%$ of the grayscale.

\subsubsection{Remarks on BM modes}

The classification of BM modes into two classes was based on a limited sample of linear operators, so it cannot be guaranteed that there are no other patterns in a larger sample of operators including non-linear operators as well.

Still subject to the same limitation, it has been found that there is no direct relationship between EQ mode and BM mode, so the two features are complementary to each other to characterize decolorization operators. For example, the standard and uniform operators have very similar BM modes, despite having EQ modes of different classes, rounded trapezoidal and bell-shaped, respectively, as shown in the figure \ref{fig:equalizer_classes}. In other words, although there are substantial differences between the number of colors that are mapped to the different gray channels, the average luminance of the colors mapped to the different gray channels by the uniform and standard operators are very similar.

On the other hand, operators with EQ modes of the same class may have BM modes of different classes. The aforementioned independence of the EQ and BM modes can be verified by analyzing, for example, the operators in the upper left and lower right corners: they have very similar EQ modes, but very dissimilar BM modes.

\subsection{Classification of Linear Color-to-Gray Operators}

Taking into account that 3 classes of EQ modes and 2 classes of BM modes were identified in linear operators, and that EQ modes and BM modes are independent of each other, it can be concluded that linear operators can be classified into 6 classes according to the EQ mode and BM mode classes.

If the two subclasses: rounded and sharp of the trapezoidal and triangular classes of EQ mode are considered, then there are 5 classes of EQ mode and 2 of BM mode, totaling 10 different classes of linear operators.

\section{Case Study}\label{sec:stydy_case}

The case study is based in a facial recognition problem. Facial recognition in images has received a lot of attention, especially for surveillance/security and social networks applications. To reduce the computational cost and storage of facial recognition systems, the images, usually in color, are stored in grayscale, so the images to be classified are previously transformed to grayscale. But many researchers have noted the impact and importance of how color transformation is done in this problem \cite{face_recon_2002,face_recon_2004, face_recon_2008, face_recon_2009,face_recon_rev_2014, face_recon_rev_2017,face_recon_rev_2019,face_recon_rev_2020,face_recon_rev_2021,face_recon_rev_2021a,face_recon_2021b,face_recon_2022}. In particular, \cite{face_recon_2004} claimed that the uniform operator was not effective for task-oriented approaches such as facial recognition. Later, \cite{face_recon_2009} came to a similar conclusion, stating that standard NTSC coefficients are not ideal for facial recognition tasks. More recently, \cite{gender_recon_2018} evaluated five gender classification algorithms developed by Amazon, Facebook, IBM, Kairos and Microsoft. They found that darker-skinned females and males, were poorly classified than lighter-skinned females and males, with error rates up to $34\%$ higher for darker-skinned female than for lighter-skinned males.

Because the cause of these errors is the imbalance of the datasets used to train the models, which predominantly contain images of Caucasian men and women, with light skin, the basic solution is to build a better balanced dataset and retrain the model. However, this is not an easy task as the robots created to automatically build the datasets, for example the used in \cite{face_dataset_2015} and \cite{face_dataset_2016}, introduce new uncertainties \cite{gender_recon_2018}.

In view of this situation, it has been hypothesized that if there was a way to lighten the face of black people in the grayscale images being processed, it might be possible to reduce the prediction error of the models trained with dataset biased towards white people. This modification could be done on the grayscale image as a preprocessing or it could be done at the time of generating the grayscale image using an appropriate operator. Opting for the second alternative, which is cheaper in terms of computation, the question arose whether it would be possible to find a linear operator that would only lighten the face without significantly reducing the contrast on the face and without significantly modifying the rest of the image. Furthermore, if this is possible, how would the EQ and BM modes of this linear operator differ compared to the uniform and standard operator?

To answer these questions, a search was carried out for the optimal linear operator for this specific task, following the procedure described in the next section.

\subsection{Experimental Procedure}

The search for an improved equalizer to lighten black people's faces was performed by the search grid method, with an independent variable. The approach taken to limit the search space was to assume one by one the $3$ $\lambda$s of the NTSC standard operator ($\lambda_r=0.299,\ \lambda_g=0.587,\ \lambda_b=0.114$) as fixed, using seven values of $K$ from $K=0.5$ to $K=12.5$ with step $2$. When $\lambda_r=0.299$ was fixed, the following version of equation \ref{eqn:1sol_quadratic} was used:
\begin{equation}
\lambda_g= \sqrt{\bigg(\frac{K\lambda_r}{2}\bigg)^2+K\lambda_r\big(1-\lambda_r\big)}-\frac{K\lambda_r}{2}
    \label{eqn:sol_quadratic_r}
\end{equation}
to calculate $\lambda_g$, and when $\lambda_g=0.587$ was fixed, the following version of equation \ref{eqn:1sol_quadratic}  was used:
\begin{equation}
\lambda_b= \sqrt{\bigg(\frac{1-\lambda_g}{2}\bigg)^2-\bigg(\frac{\lambda_g^2}{K}\bigg)}+\frac{(1-\lambda_g)^2}{2}
    \label{eqn:sol_quadratic_g}
\end{equation}
to calculate $\lambda_b$. This equation is valid for $K\ge 8.08$, and then  only three $K$ values ($8.5,10.5$ and $12.5$) were considered in this case. Thus, $17$ operator configurations were analyzed.

The dataset was comprised of $46$ test face images, $24$ of black people and $22$ of white people. The $17$ grayscale versions of the $24$ images of black faces were evaluated by $12$ observers, who chose the one they considered easiest to identify, without knowing which setting was used. The $17$ grayscale versions of each color test image were presented for the first time in a $3 x 6$ mosaic with random placement. In one corner of the screen, the original color image was shown for reference. In this mosaic, the viewer chose the four best ones, which were then presented in a $2 x 2$ mosaic on the next screen asking to choose the best one. At this step, the observer could view one of the images enlarged for detailing. Of the $12 \times 24 = 288$ decisions, $153$ favored the chosen operator. After choosing the most voted configuration with blacks, the images of the white faces were presented and the process was repeated. Of the total of $12 \times 22 = 264$ decisions, $132$ votes favored the same configuration chosen for black faces. The second most voted operator received only $36$ votes. 

\section{Results and Discussion}\label{sec:res_disc}

\begin{figure}
    \centering
        \includegraphics[scale=0.3]{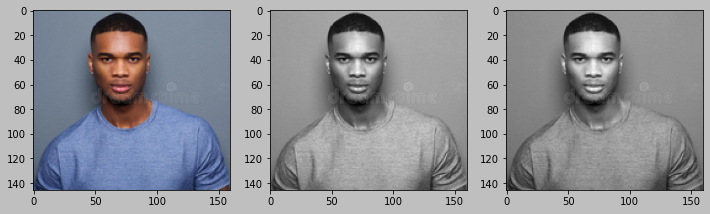}
        \includegraphics[scale=0.3]{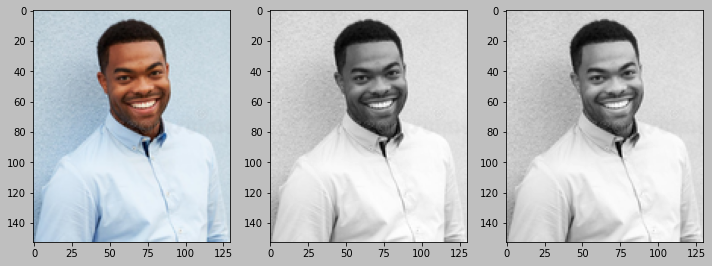}
        \includegraphics[scale=0.3]{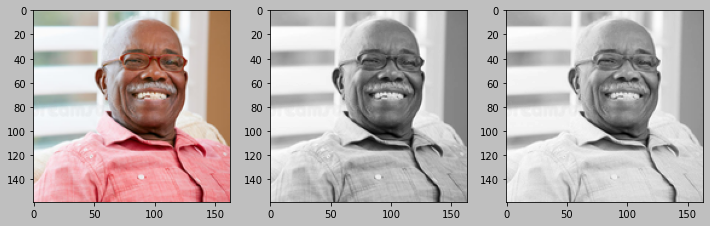}
        \includegraphics[scale=0.3]{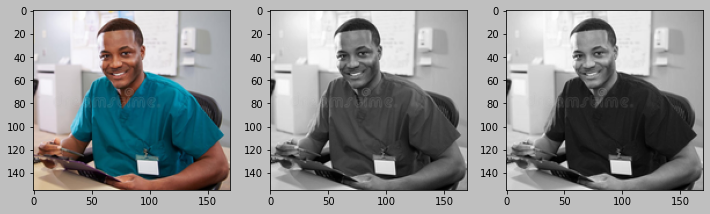}
        \includegraphics[scale=0.3]{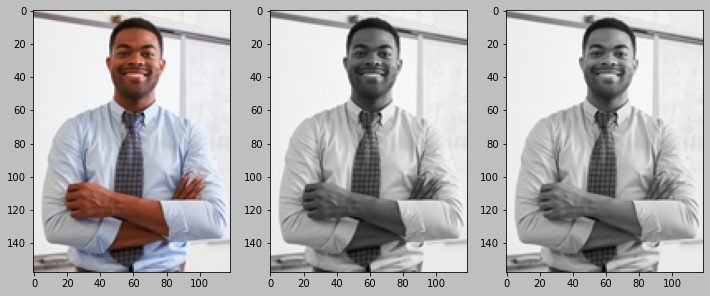}
        \includegraphics[scale=0.3]{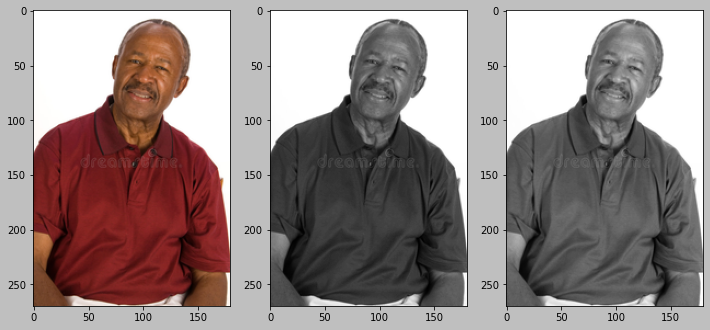}
    \caption{Black man faces transformed with the standard (center) and chosen as best (right) operators. Source: dreamstime's public repository.}
    \label{fig:mb}
\end{figure}

\begin{figure}
    \centering
        \includegraphics[scale=0.3]{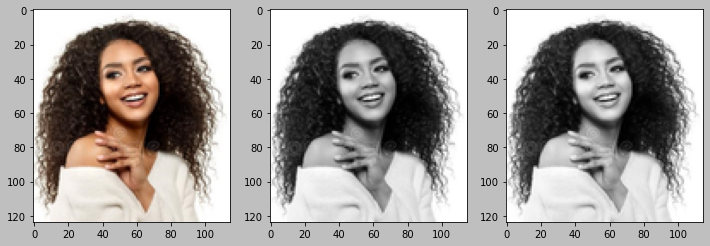}
        \includegraphics[scale=0.3]{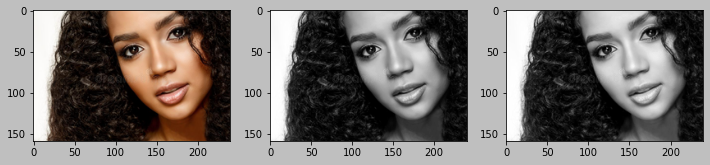}
        \includegraphics[scale=0.3]{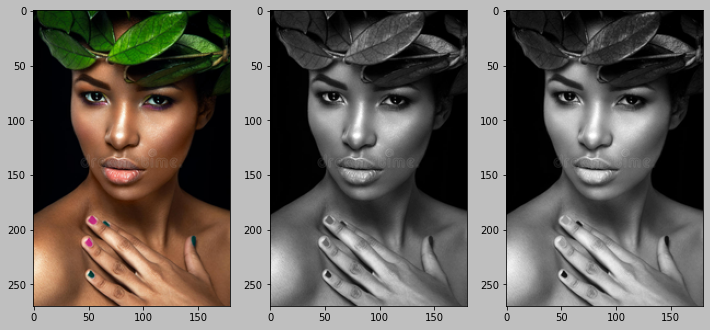}
        \includegraphics[scale=0.3]{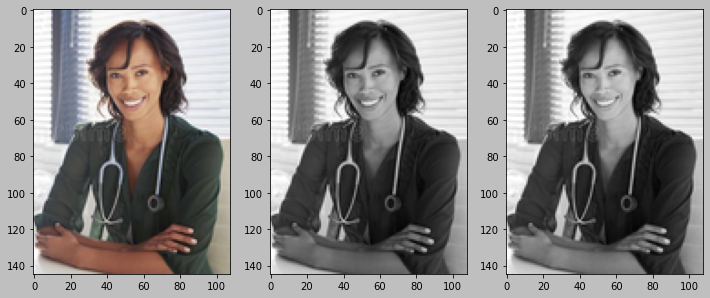}
        \includegraphics[scale=0.3]{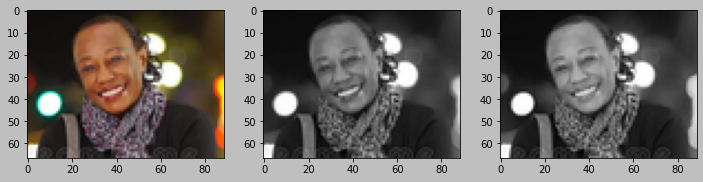}
        \includegraphics[scale=0.3]{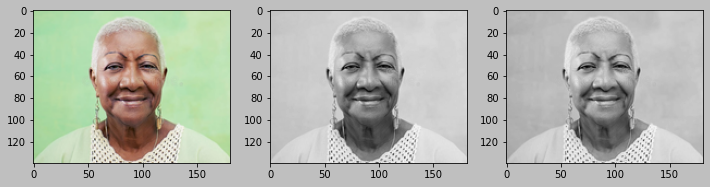}
    \caption{Black woman faces transformed with the standard (center) and optimized (right) operators. Source: dreamstime's public repository.}
    \label{fig:wb}
\end{figure}

The differentiated effect of the selected (optimal) operator  transforming images of black people, can be seen in the figures \ref{fig:mb} and \ref{fig:wb}. It shows a mosaic with 6 of the faces of black men and women, respectively, considered in the experiments, transformed with the standard (center) and optimized (right) operators. 
It can be noticed that the skin of the face is lightened with the chosen equalizer without significantly affecting other features of the face such as hair, mouth, eyes and nose, nor the background. These results respond positively to the question of whether it would be possible to find a linear operator that would only lighten the face without significantly affecting the other properties and regions of the image. 

\begin{figure}
    \centering
        \includegraphics[scale=0.3]{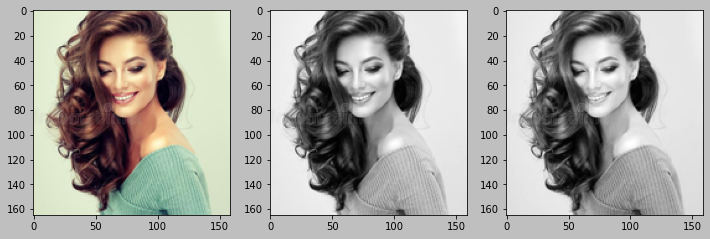}
        \includegraphics[scale=0.3]{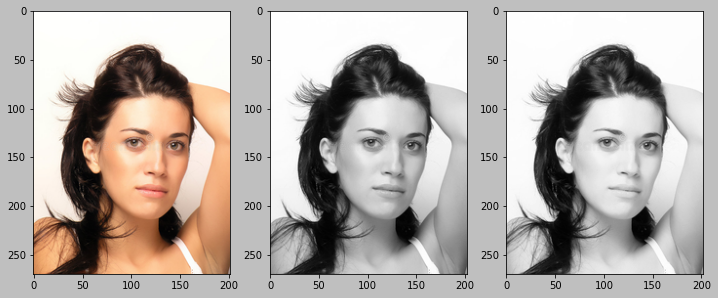}
        \includegraphics[scale=0.3]{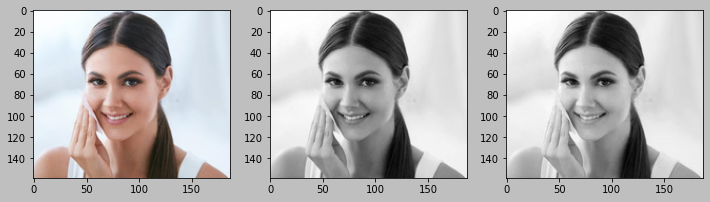}
        \includegraphics[scale=0.3]{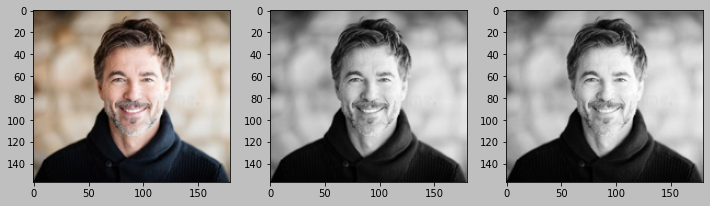}
        \includegraphics[scale=0.3]{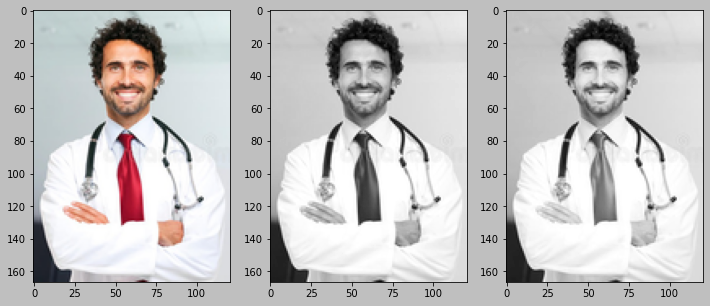}
        \includegraphics[scale=0.3]{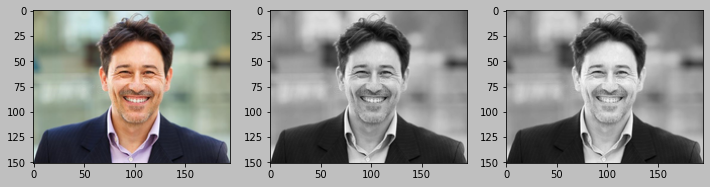}
    \caption{White woman and man faces transformed with the standard (center) and optimized (right) operators. Source: dreamstime's public repository.}
    \label{fig:wmw}
\end{figure}

The selected operator configuration is: $K=0.5$, $\lambda_b=0.114$, $\lambda_g=0.198$ and $\lambda_r=0.688$, showing a predominance of the red channel, which corresponds to that reported in \cite {face_recon_2009} that increased the contribution of the red channel to improve the performance of the face identification algorithm based on the Adaboost method. An interesting fact is that this equalizer also lit up the whites' faces, as can be seen in the figure \ref{fig:wmw}.

This result shows that the starting hypothesis is valid, as it has been possible to find a linear operator capable of meeting the needs of the task, although quite peculiar, in the sense of producing an effect located in a specific region of the images: on people's faces.

Without the intention of generalizing, the result shown suggests that it is worth exploring the space of linear operators looking for one that meets specific requirements and only in case of failure to move on to another more complex solution.

\begin{figure}[h!]
    \centering
    \includegraphics[scale=0.4]{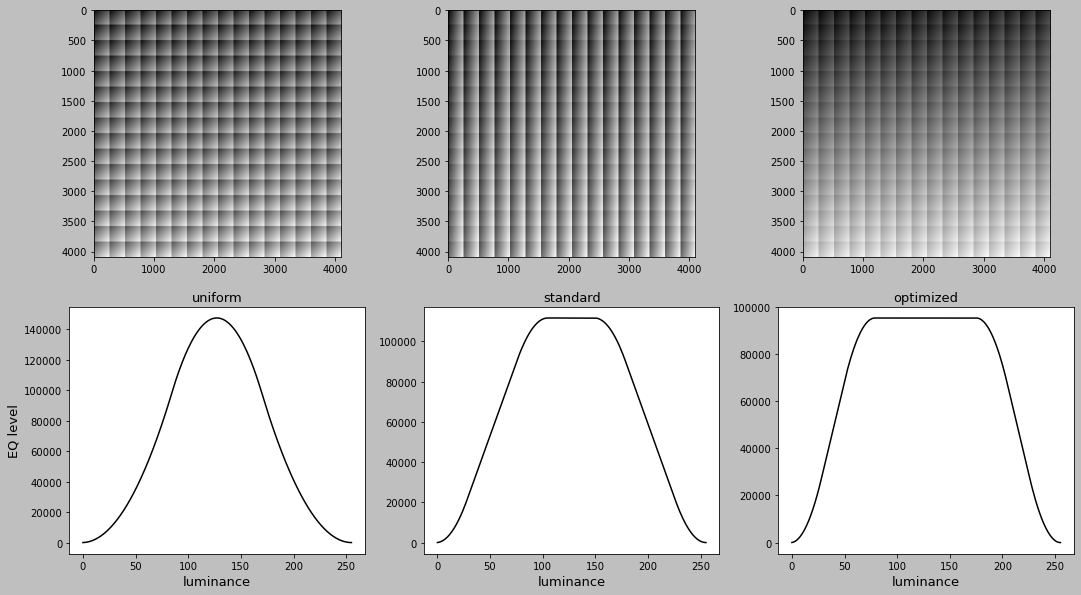}
    \caption{Equalization modes of the uniform, standard and optimized operators and the corresponding grayscale versions of the 2D reference image.}
    \label{fig:ref_2Dimg_to_rgb_3_EQ}
\end{figure}

Moving on to the comparison of the standard operator with the selected one (optimal for this task), figure \ref{fig:ref_2Dimg_to_rgb_3_EQ} shows the EQ modes of the uniform, standard and chosen operators. It also shows the grayscale images converted from the 2D reference image.
The first observation is that the optimal operator belongs to the same equalizer class as the standard operator: rounded trapezoidal, however, the plateau region is much wider.

Comparing the gray versions, it has been observed that the chosen operator produces a greater uniformity inside the boxes, increasing the total brightness of the boxes in the downward direction, creating a macro gradient in the image. In the uniform and standard operators, the macro gradient of  brightness is not so well perceptible, but the micro gradients (inside the boxes) are more intense. In visual terms, the image produced with the optimized operator presents the greatest global contrast in the 2D grayscale image, while the uniform operator presents the lowest global contrast.

\begin{figure}[h!]
    \centering
    \includegraphics[scale=0.5]{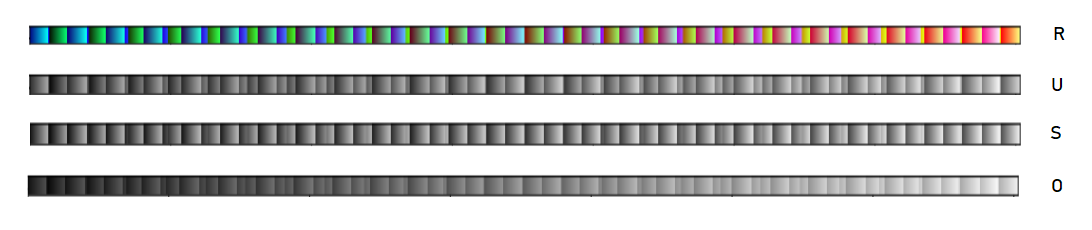}
    \caption{Effect of the EQ mode of the uniform (U),  standard (S) and optimized (O) operators and the 1D reference image (R).}
    \label{fig:ref_1Dimg_to_rgb_3_EQ}
\end{figure}

Figure \ref{fig:ref_1Dimg_to_rgb_3_EQ} shows the 1D reference image and the 1D grayscale images produced with each operator. This type of visualization is much more compact and carries the same information as the 2D image. Notice that there is a brightness macrogradient to the left, with the optimized operator having the smoothest macrogradient.

\begin{figure}[h!]
    \centering
    \includegraphics[scale=0.35]{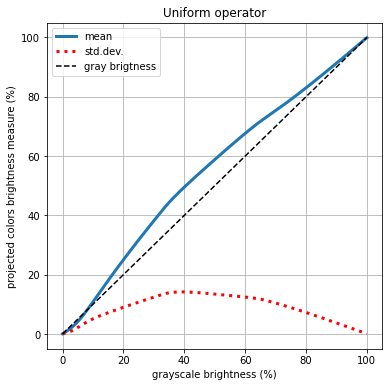}
    \includegraphics[scale=0.35]{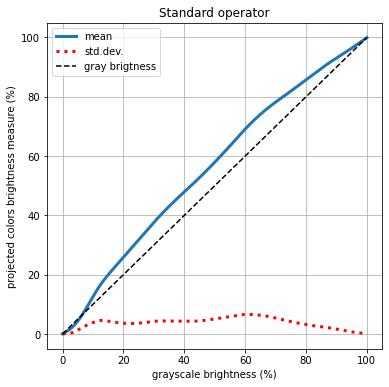}
    \includegraphics[scale=0.35]{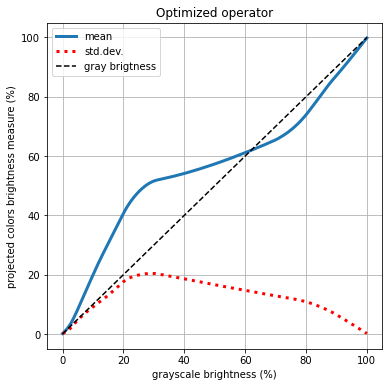}
    \caption{BM modes of the uniform, standard and optimized operators (blue line). Black dashed line represents the grayscale brightness of the projected colors (function $y=x$). Red dotted line is the standard deviation.}
    \label{fig:BMM_3operators}
\end{figure}

Regarding the BM modes, figure \ref{fig:BMM_3operators} shows the results for the uniform, standard and chosen operators. 
It can be seen that the BM mode of the optimized face lightening operator belongs to a different class than the other two operators. The chosen operator, in approximately $2/3$ of the luminance range is above but in the final part of the range is below the iso-brightness line, which indicates that for the high-brightness regions of the grayscale, not-so-bright colors are projected. This inversion can be an experimental evidence of the cause of the observed facial whitening effect.

\section{Conclusions}\label{sec:concl}

At the conclusion of the article, it is worth summarizing the contributions of this work:

\begin{enumerate}
   \item The introduction of a synthetic reference image, which is an essential resource for the analysis of projection operators of any type, not just the linear ones considered in this work. The reference image is composed of a single pixel of each of the $256^3$ colors, ordered in a specific way, chosen arbitrarily. Consequently, two algorithms to generate the 1D (vector) and 2D (matrix) versions of the reference image were described to allow its faithful reproduction, maintaining the original order of the colors.

    \item Based on the association of color-to-gray conversions with a task of grouping colors into $256$ clusters with increasing gray brightness, two features to characterize the color-to-gray projection operators were introduced:

     \begin{enumerate}
         \item The Equalization mode (EQ mode), which describes the size of the clusters, and
         \item The Brightness Mapping mode (BM mode), which describes the average brightness of the clusters.
     \end{enumerate}
     
    \item A taxonomy of color-to-gray linear operators has been defined: It has been shown that the infinite population of color-to-gray linear projection operators can be classified into only six classes considering the three EQ mode classes (bell shaped, triangular and trapezoidal) and the two BM mode classes (regular and irregular). If the two subclasses of the triangular and trapezoidal EQ modes, that is, the rounded and the sharp versions, are considered,  then the linear operators can be classified into ten classes. This taxonomy can be extended by including the EQ/BM mode patterns of non-linear operators. 
    
    \item A parametric classification has been introduced that groups linear operators into families, as well as assigns a unique ID to each operator instance within its family.
    
     \item A linear operator for converting color to gray has been defined that has a face lightening effect, whose weights are: $\lambda_r=0.688$, $\lambda_g=0.198$ and $\lambda_b=0.114$.

    \item It was shown with a concrete practical example that it is necessary to study further the coherence between the error metrics used to evaluate the quality of conversions from color images to grayscale and the subjective perceptual evaluation, since there was a substantial discrepancy in the case study. 
    
    \item Finally, it has been shown that even a simple linear operator can achieve interesting and useful effects not only globally, as intuitively expected, but also locally.
    
\end{enumerate}
 
 The analytical approach from the new perspective described in this work can be useful for a better understanding of the current state-of-the-art color-to-gray operators.
 
\bibliographystyle{alpha}
\bibliography{main}

\end{document}